\documentclass[a4paper,11pt]{article}

\usepackage{natbib}
\usepackage[a4paper,margin=3cm]{geometry}
\newcommand{\cites}[1]{\citeauthor{#1}'s \citeyearpar{#1}}
\newcommand{\citeasnoun}[1]{\citet{#1}}
\renewcommand{\cite}[1]{\citep{#1}}
\usepackage{tikz}
\usetikzlibrary{arrows,matrix,graphs,shapes,snakes,automata,backgrounds,petri,positioning,calc,decorations.pathmorphing}
\usepackage[leqno]{amsmath}
\usepackage{amssymb}
\usepackage{latexsym}
\usepackage{amsfonts}
\usepackage{proof}
\usepackage{linguex}
\usepackage{xcolor}

\colorlet{lightgray}{gray!10}
\newcommand{\myindex}[1]{\!\!\footnotesize #1\!\!}
\newcommand{\myindexi}[1]{\!\!\footnotesize \textit{#1}\!\!}
\newcommand{\apsnodei}{\centerdot_{\rule{0pt}{1.2ex}}}
\newcommand{\ldr}{\mathbin{/}}
\newcommand{\ldl}{\mathbin{\backslash}}
\newcommand{\lpr}{\bullet}

\newcommand{\bridge}{\,\hat{\,}}

\newcommand{\downr}{\downarrow_{<}}
\newcommand{\upl}{\uparrow_{>}}
\newcommand{\upr}{\uparrow_{<}}

\newcommand{\editout}[1]{}

\tikzstyle{reverseclip}=[insert path={(current page.north east) --
  (current page.south east) --
  (current page.south west) --
  (current page.north west) --
  (current page.north east)}
]

\tikzset{pas/.style={fill=gray!60}, 
act/.style={fill=gray!30},
main/.style={draw,fill=white},
ctx/.style={rounded rectangle,minimum size=7mm},
val/.style={rectangle,minimum size=7mm},
cmd/.style={chamfered rectangle,draw,fill=white},
tns/.style={circle,minimum size=3mm,draw,fill=white},
par/.style={circle,minimum size=3mm,draw,fill=black}, 
minipar/.style={circle,minimum size=2.5mm,draw,fill=black}, 
pn/.style={rounded corners, rectangle,fill=blue!30,draw,minimum size=15mm},
medpn/.style={rounded corners, rectangle,fill=blue!30,draw,minimum size=20mm},
 bigpn/.style={rounded corners, rectangle,fill=blue!30,draw,minimum size=25mm}}

\title{Comparing and evaluating\\ extended Lambek calculi}
\author{Richard Moot\thanks{This work has benefitted from the generous
    support of the French agency Agence Nationale de la Recherche as
    part of the research project Polymnie (ANR-12-CORD-0004)} \\
CNRS (LaBRI) \\ University of Bordeaux}

\begin{document}

\maketitle

\thispagestyle{empty}

\begin{abstract}
Lambek’s Syntactic Calculus, commonly referred to as the Lambek calculus, was innovative in many ways, notably as a precursor of linear logic. But it also showed that we could treat our grammatical framework as a logic (as opposed to a logical theory). However, though it was successful in giving at least a basic treatment of many linguistic phenomena, it was also clear that a slightly more expressive logical calculus was needed for many other cases. Therefore, many extensions and variants of the Lambek calculus have been proposed, since the eighties and up until the present day. As a result, there
is now a large class of calculi, each with its own empirical successes
and theoretical results, but also each with its own logical primitives. This raises the question: how do we compare
and evaluate these different logical formalisms?

To answer this question, I present two unifying frameworks for these extended
Lambek calculi. Both are proof net calculi with graph contraction
criteria. 

The first calculus is a very general system: you specify
the structure of your sequents and it gives you the connectives and
contractions which correspond to it. The calculus can be extended with
structural rules, which translate directly into graph rewrite rules.

The second calculus is first-order (multiplicative intuitionistic)
linear logic, which turns out to have several other, independently
proposed extensions of the Lambek calculus as fragments.

I will illustrate the use of each calculus in building bridges between
analyses proposed in different frameworks, in highlighting
differences and in helping to identify problems.
\end{abstract}

\section{Introduction}

The Lambek calculus \cite{lambek} is a landmark formal system. It is a logically
simple system, with a transparent interface to natural language
semantics \cite{cglambda} and gives a basic account of some interesting phenomena on the
syntax-semantics interface, such as quantifiers and extraction. However, researchers working on the Lambek calculus soon suspected that
there were problems with the calculus, both of a formal and of a
descriptive nature.

Though this was only proved in \cite{pentus}, it has long been suspected
that Lambek grammars generate only context-free languages and there
are compelling arguments that
natural languages require at least a slightly larger class of
languages \cite{shieber,tagcs}.

On the descriptive side, the Lambek calculus handles quantifiers and extraction only
from peripheral positions: we can approximate medial extraction and
medial quantifiers taking wide scope only
by multiplying and complicating the lexical entries. Many other
phenomena are problematic for the Lambek calculus.

Besides these problems of undergeneration, it has been known since
\cite{lambek61} that the Lambek calculus suffers from overgeneration
as well. Lambek lists several problem cases, but the global availability of the associativity rule means that
we predict very bad coordinations such as ``*The
mother of and John thinks that Bill left''\footnote{Paul Dekker was
  the first to note this example which has been mentioned in the
  literature since the early 90s. Under global associativity and
  standard lexical assignments, both ``the mother of'' and ``John
  thinks that'' are expressions of type $(s\ldr (np\ldl s))\ldr np$.}. Lambek's (1961)
non-associative Lambek calculus, NL, solves these problems of
overgeneration but in doing so makes the system too restrictive in
other ways. For example the simple treatment of peripheral extraction
and quantification of the Lambek calculus is no longer available in NL.

One of the main research goals in type-logical grammars has been to
keep the good properties of the Lambek calculus (notably its logical
simplicity and its connection to natural language semantics) while
solving its empirical problems accounting for linguistic data.

In pursuit of this goal, a rather large number of logics has been proposed, which
include the Lambek-Grishin calculus (LG) \cite{bernardi10conti}, the
Displacement calculus \cite{mvf11displacement}, multimodal categorial
grammars \cite{mmli},
lambda grammars \cite{oehrle}, $\textrm{NL}_{\lambda}$ \cite{bs14cont}, Hybrid Type-Logical
Grammars \cite{kl12gap}, etc.\ All of these logics are a shift of perspective with
respect to the original Lambek calculus: for example, the Displacement
calculus changes the basic objects from strings to string tuples
(thereby making it natural to add logical operations for infixes and
circumfixes), whereas Hybrid Type-Logical Grammars add the lambda
calculus term constructors of abstraction and application (and the
associated reduction operations)
to the Lambek calculus connectives.

There is a
rather large number of logical calculi available all claiming to
improve upon the Lambek calculus in one way or another. However, this
proliferation of calculi makes it hard to compare and evaluate the
benefits of different calculi with respect to one another. 

Given the variety of logical primitives used in these different logics, we should not expect a single ``silver bullet'' which relates all these different calculi. 
However, I discuss two ``meta-logics'' which at least provide
two general strategies for doing such comparisons and give some examples of how to apply them. 

The rest of the paper in structured as follows. I will first briefly
discuss the common core shared by type-logical grammars:
multiplicative intuitionistic linear logic as the syntax-semantics
interface, followed by a very brief discussion of some of the phenomena in syntax
and the syntax-semantics interface which we want our
type-logical grammars to treat.

Then, I will present two meta-logics both employing a variant of
\cites{reductions} graph contractions. The first system is a general
and flexible proof net system which can be adapted to different
connectives and different structural rules. The second system is
first-order linear logic and it has several other type-logical
grammars as a special case.

In what follows, I will assume the reader is familiar with the Lambek calculus and has
some basic knowledge of formal semantics and of proof theory.

\section{Multiplicative intuitionistic linear logic and the
  syntax-semantics interface}
\label{sec:ss}

One standard architectural choice is shared between the different
type-logical grammars discussed in this paper: the syntax-semantics
interface takes the form of a homomorphism from proofs in the given
source logic to proofs in multiplicative intuitionistic linear
logic\footnote{The alternative is to use a simple, applicative logic
  and to add a set of type-shifting rules in the spirit of combinatorial logic.
Two types categorial grammars using this alternative setup are
combinatory categorial grammar \citep[CCG]{steedman} and flexible
Montague grammar \cite{hh93sf}. The choice of multiplicative
intuitionistic linear logic also has some descriptive repercussions:
as discussed in Section~\ref{sec:cat}, the treatment of parasitic gapping
and some treatments of anaphora and binding are incompatible with this
choice and require a more powerful logic.}
(also called the Lambek-van Benthem calculus LP, in the categorial
grammar literature). 

In what follows, we will call the multiplicative intuitionistic linear
logic proof the \emph{deep structure} of a sentence and the proof in
the source logic the \emph{surface structure}. From the deep
structure we obtain the meaning of a sentence by substituting the
semantic terms from the lexicon and normalising the result term.

Seen from this perspective, different type-logical grammars generally
agree on the deep structure for an analysis, 
  but arrive at this deep
structure from different surface structures. As \citet[Section~3]{M95} remarks, the tension between the LP proofs of deep
structure and the more restricted proofs of surface structure has
played an important role in the development of type-logical
grammars. We have to find a delicate balance between avoiding overgeneration and
generating all the desired readings of our sentences.

\section{A catalogue of problem cases}
\label{sec:cat}

What are the linguistic data which motivated these new calculi? The
following list gives some of the types of phenomena people have look
at. Since a finite list of cases can always be treated by some
additional lexical type assignments, we will be interested
only in \emph{robust} solutions, that is solutions which generalise
beyond the listed examples to more complex cases, and ideally to
different \emph{phenomena}. In other words, we want to avoid ad hoc
solutions (though, as an illustration, I will
sometimes give an additional assignment which
solves a particular case).

As usual, an asterisk ``*'' before a sentence denotes
ungrammaticality.


\ex.\label{rnr} John loved but Peter hated ``Syntactic Structures''.

\ex. \label{island} *Peter bough the book which John read ``Syntactic
Structures'' and Mindy liked.

\ex.\label{medex} Peter bough the book which John read yesterday.

\ex.\label{medq}  John believes someone left.

\ex.\label{dutch} (dat) Jan Henk Marie de nijlpaarden zag helpen voeren.

\ex.\label{gap}  John studies logic and Charles, phonetics.

\ex.\label{ellipsis} John left before Mary did.

\ex.\label{compa} Mary gave more girls a book than boys a record.


Example~\ref{rnr}, called ``right node raising'' in the literature, shows one of the
things which works correctly in the Lambek calculus: assuming
assignments of $(np\ldl s)\ldr np$ to the transitive verbs and $np$ to
``John'' and ``Peter'', L derives ``John
loved'' and ``Peter hated'' as expressions of type $s\ldr np$. Though
this works correctly, it crucially depends on the presence of
associativity. For example, NL no longer allows us to derive
sentence~\ref{rnr} unless we add a second lexical assignment $np\ldl
(s \ldr np)$ to the transitive verbs.

Examples like~\ref{island}, and related examples, were first discussed
in \citep[p.\ 167]{lambek61}. The problem with this ungrammatical
sentence is that ``John read \emph{Syntactic Structures} and Mindy
liked'' is a sentence missing a noun phrase (at its right edge) and
can therefore combine with ``which'' given its standard Lambek
calculus assignment $(n\ldl n)\ldr (s\ldr np)$. The same derivability
pattern which helped us for sentence~\ref{rnr} leads to overgeneration
for sentence~\ref{island}. The ungrammaticality of this sentence is
usually ascribed to ``island constraints''.

Examples~\ref{medex} and~\ref{medq} illustrate that the Lambek
calculus has problems with quantifiers in medial position taking wide
scope (``someone'' in the \emph{de re} reading of example~\ref{medq})
and extraction from non-peripheral positions: sentence~\ref{medex} is
derivable without the sentence-final adverb ``yesterday'', again since
``John read'' is an expression of type $s\ldr np$. However, the
presence of ``yesterday'' blocks this derivation.

Another classic example is the treatment of Dutch verb clusters
\cite{MO94,oehrle11multi} (Sentence~\ref{dutch} above). This sentence exhibits
the well-known crossed dependencies: ``Henk'' is the object of
``zag'' (saw), ``Marie'' the object of ``helpen'' (help) and ``de nijlpaarden''
(the hippopotami) the object of ``voeren'' (feed). Though
\citeasnoun{gp82cfg} show that such examples can be treated by
context-free grammars, and hence by the Lambek calculus, the Dutch
verb cluster analysis
of mildly context-sensitive formalisms, which expresses the desired
dependencies between objects and verbs is generally preferred.

Gapping \cite{cgellipsis}
(Sentence~\ref{gap}, which has the same meaning as ``John studies
logic and Charles \emph{studies} phonetics''), ellipsis
(Sentence~\ref{ellipsis}, which has the same meaning as ``John left
before Mary \emph{left}'') and comparative subdeletion
(Sentence~\ref{compa}, which means ``The number of girls Mary gave a
book is greater than the number of boys \emph{Mary gave} a record''),
are some of the other interesting phenomena on the syntax-semantics interface
which have been treated in type-logical grammars.

One general type of example where type-logical grammars do well are
coordination and similar phenomena. The type-logical grammar treatment
of coordination tells us we can conjoin two phrases whenever they can
be assigned the same formula. Sentence~\ref{rnr} is an example, but
also, in the analysis of \citeasnoun{cgellipsis},
Sentence~\ref{gap}\footnote{Though in the case of \cite{cgellipsis} and
  also in the similar case of \cite{mvf11displacement} this is a
  coordination of two formulas which are \textit{almost} the same.}. The easiest way to implement this is to assign the
second-order formula $\forall X. (( X\ldl X)\ldr X)$ to words like
``and'' and ``but'', though how to add second-order quantifiers while
keeping type-logical grammars decidable is an open question.

A general type of extension to the Lambek calculus is exemplified by
\cites{m96q} $q(A,B,C)$ constructor, which has a rather simple left
rule.

$$
\infer[qL]{\Gamma[\Delta[q(A,B,C)]]\vdash D}{\Delta[A]\vdash B &
  \Gamma[C]\vdash D}
$$

Its instantiation $q(np,s,s)$ gives a good account of quantifier
scope, whereas reflexives can be assigned $q(np,np\ldl s,np\ldl s)$. In addition, it can be extended
to treat comparative subdeletion and other phenomena as well
(Sentence~\ref{compa}) \cite{m96q}. \cites{TLS} textbook gives an introduction to semantics and
type-logical grammar using
mostly the Lambek calculus extended with the $q$ operator.
Now, semantically, the $q$ operator behaves like a combination of an introduction rule
and an elimination rule, so it seems the $q$ connective may need to be
decomposed and though we have given a left rule for $q$ there is no
easy way to formulate its \textit{right} rule for $q$. Because of the 
simplicity and many applications of the $q$ operator, these questions
have led to many decompositions of $q$ \citep[to cite but a few]{morrill,mac10,bs14cont}

The above list of sentences gives only a very partial
picture of the descriptive work done in the various type-logical
frameworks, but it provides a starting point for discussion.

\begin{table}
\begin{center}
\begin{tabular}{lccccccllcccccc}
Formalism & Islands & RNR & Rel & $q$ & $\wedge$ & gap & Language & Complexity\\
NL             & +          & --     & --    & --  &  +             &
                                                                      --
                                                       &= CFL & P \\
L               & --           & +     & --    & --  &  +& --& = CFL &
  NP\\
D$_{\textit{core}}$ & -- & + & + & + & + & + & $\supseteq$ MCFL$_{wn}$
               & NP\\ 
D$_{\textit{full}}$  &+ & + & + & + & + & + & $\supsetneq$ MCFL & ?\\  
ACG$_2$   & --           & --     & --    & --  & -- & -- & = MCFL &NP\\
ACG  & -- & -- & + & (+) & -- & -- & $\supseteq$ MCFL &NP\\
LG & -- & -- & -- & + & + & + & $\supsetneq$ MCFL &NP\\
NL$_{\lambda}$ & + & -- & + & + & + & + & $\supsetneq$ MCFL &NP(?)\\
HTLG & -- & + & + & + & + & + & $\supsetneq$ MCFL &NP\\
MILL1 & + & + & + & + & + & + & $\supsetneq$ MCFL &NP\\
NL$\Diamond_{\mathcal{R}}$ & + & + & + & + & + & + & = CSL &PSPACE\\
\end{tabular}
\end{center}
\caption{Scorecard for the different extended Lambek calculi}
\label{tab:score}
\end{table}


Table~\ref{tab:score} is my attempt to, as \citeasnoun{lakatosappraising} suggests, keep score
honestly. It is a table cross-referencing formalisms and phenomena,
assigning a ``$+$'' if the formalism has a good treatment of the given
empirical data and ``$-$'' if it does not.

The formalisms are, from top to bottom: the non-associative Lambek
calculus, NL \cite{lambek61}, the associative Lambek calculus L
\cite{lambek}, the core, multiplicative fragment of the Displacement
calculus, D$_{\textit{core}}$ \citep[this includes the multiplicative
connectives, with the synthetic connectives restricted to $\bridge$,  $\triangleright^{-1}$ and
$\triangleleft^{-1}$, when I use the term ``D'' or ``the Displacement
calculus'' without further qualification in the rest of this paper, I
mean this fragment]{mvf11displacement}, the full Displacement
calculus, D$_{\textit{full}}$ \citep[this volume]{mv15d}. Abstract
categorial grammars/lambda grammars, ACG and their second-order restriction ACG$_2$
\cite{groote01acg,degroote-pogodalla05}, the Lambek-Grishin calculus,
LG \cite{bernardi10conti}, NL$_{\lambda}$ \cite{bs14cont}, Hybrid
Type-Logical Grammars, HTLG \cite{kl12gap,kl13dgap}, first-order linear
logic, MILL1 \cite{mill1} and the multimodal Lambek calculus
NL$\Diamond_{\mathcal{R}}$ \cite{mmli,M95}.

The phenomena are, from left to right: does the formalism have a way,
such a non-associativity, to encode island constraints? Does the
formalism have a way to encode right-node raising (RNR)? Does the
formalism have a way to handle medial extraction? Does the formalism
have a way to simulate the $q$ operator\footnote{A ``(+)'' means the
formalism can only do so partially. For example ACG does not handle $q(A,B,C)$ for
complex formulas, for example those used for reflexives
\cite{hal-00996724}.}? Does the formalism handle
coordination ($\wedge$)\footnote{Coordination for atomic formulas only is marked as
``$-$'', a ``$+$'' for coordination means: ``the formula $\forall
X. (( X\ldl X)\ldr X)$  for conjunction works correctly for all
formulas $X$ in our lexicon''}? Does the formalism treat gapping and related
phenomena? What do we know about the language class of
the given formalism (the classes are context-free languages,
well-nested multiple context-free languages, multiple context-free
languages and context-sensitive languages)? And finally, what is the
computational complexity of the formal system?

Our knowledge of the precise language classes is very partial: this is
partly due to the difficulty of extending a proof like \cite{pentus}
to extended Lambek calculi \citep{Bus97}, which means few upper bounds
are known and in general few formal proofs of generated language
classes exist and the table entries represent answers to questions
like ``are there grammars of the language $a^nb^nc^n$,  and
$ww$?''. Formalisms which generate (at least) the well-nested
multiple context-free languages can correctly treat phenomena such as
the Dutch verb
clusters discussed above.

Not listed in the table is parasitic gapping.
The full Displacement calculus with
contraction modalities is the only calculus among those listed
which has a treatment of pa\-ra\-si\-tic gapping, as exemplified by a noun
such as.

\ex. contract which John filed without reading.

In these cases, the object noun phrase bound by ``which'' has been
extracted twice, once as the object of ``filed'' and once as the
object of ``reading''.

Also absent from the table are treatments of anaphora rooted in
logical syntax, such as those of
\cite{jaeger,mv14d,bs14cont}. Like parasitic gapping, such treatments
require a modification of the hypothesis, discussed in Section~\ref{sec:ss}, of a linear syntax-semantics
interface and like parasitic  gapping, they complicate the proof
theory. I believe that for anaphora (and binding theory) the proper
division of labour between syntax, syntax-semantics interface and
semantics/pragmatics is still very much open for debate and that
integrating the many interacting facts on these different levels into
a single, coherent system will be a difficult but important task.

The computational complexity column lists
the complexity of the parsing or universal recognition problem (fixed
recognition is a property of formal languages and has little relation
to parsing complexity). All results
listed ``NP'' or ``PSPACE''
indicate NP- or PSPACE-completeness. The complexity of the
full Displacement calculus is unknown, and may be
undecidable. The complexity of NL$_{\lambda}$ is unknown;
\citet{bs14cont} show only decidability, but the close connection to
HTLG --- both employ a combination of a standard Lambek calculus (L
and NL respectively) together with lambda-calculus like operations
(exactly those of the lambda calculus for HTLG and operations very
close to it in NL$_{\lambda}$) --- makes it likely to be an NP-complete system (since it appears
polynomial time verifiable whether or not an HTLG proof is also an
NL$_{\lambda}$ proof\footnote{Of course this presupposes first that
  NL$_{\lambda}$ proofs are a subset of the HTLG proofs, in the same
  sense that NL proofs are a subset of L proofs.}).
NL can be parsed in polynomial time \cite{aarts95,nlpoly}.

Though it is easy to criticise this table and talk
about important missing phenomena or incorrectly assigned entries, I
hope this will provide a starting point for discussion and provide an
impetus for people to ``improve their score'' on the table. I would
like to add that systems with a ``low score'', such as the Lambek
calculus, are still worthwhile systems to use and study, as long as we are aware
of their limitations for certain applications.
My goal is to allow ``people to do their own thing but only as long as
 they publicly admit what the score is between them and their rivals''
 \citep[p.\ 100]{lakatosappraising}.
\section{Graph rewriting}
\label{sec:graph}

The first meta-logic for creating bridges between the different
syntactic derivations of extended Lambek calculi
 is a simple generalisation of the proof net calculus of
\citeasnoun{mp}. Its basic idea is that you only need to specify the
\emph{structures} of your logic (ie.\ the way your formulas are
organised in the sequence, for example, as a list, a tree, or as a
graph) and then you will automatically obtain the possible connectives and a correctness criterion
based on graph rewriting and contractions in the style of
\citeasnoun{reductions} and close to the interaction nets of
\cite{lafont95}. 

For example, if we choose binary branching trees as structures, then
we obtain three connectives depending on which of the three related
nodes contains the main formula and these are just the familiar
connectives of the non-associative Lambek calculus NL.

\begin{figure}
\begin{center}
\begin{tikzpicture}[node distance=3mm]
\node[tns] (c) at (-6em,2em) {};
\node (b) at (-6em,0em) {$C$};
\node (l) at (-8em,3.6em) {$C / B$};
\node (r) at (-4em,3.6em) {$B$};
\draw (b) -- (c);
\draw (c) -- (l);
\draw (c) -- (r);
\node[tns] (c) at (2em,2em) {};
\node (b) at (2em,0em) {$A\bullet B$};
\node (l) at (0em,3.6em) {$A$};
\node (r) at (4em,3.6em) {$B$};
\draw (b) -- (c);
\draw (c) -- (l);
\draw (c) -- (r);
\node[tns] (c) at (10em,2em) {};
\node (b) at (10em,0em) {$C$};
\node (l) at (8em,3.6em) {$A$};
\node (r) at (12em,3.6em) {$A\backslash C$};
\draw (b) -- (c);
\draw (c) -- (l);
\draw (c) -- (r);
\end{tikzpicture}
\end{center}
\caption{Binary branches and the connectives of NL}
\label{fig:nll}
\end{figure}
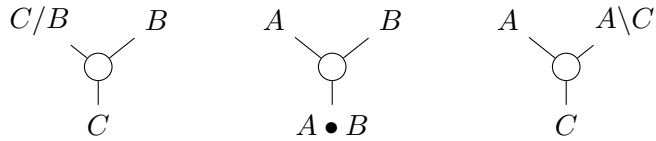

In addition to the constructors shown in Figure~\ref{fig:nll}, each connective, by
logical symmetry, induces a destructor\footnote{For the binary
  connectives of linear logic, the constructors are tensor links and
  the destructors are par links. In the terminology of focusing proof
  search \citep{focus}, we would talk about synchronous or non-invertible rules instead of
constructors and asynchronous or invertible rules instead of destructors.} as well. These destructors
have the property that an appropriately connected
constructor/destructor pair contracts to a point. When it is
impossible to contract a destructor, the graph does not correspond to
a provable statement; provable statements contract to a (constructor) tree. 

\begin{figure}
\begin{center}
\begin{tikzpicture}[scale=0.75]
\node (labl) at (3em,-2.5em) {$[R\ldr ]$};
\node (ab) at (3em,4.8em) {$\apsnodei$};
\node (a) at (0,9.6em) {$\apsnodei$};
\node (b) at (6em,9.6em) {$\apsnodei$};
\node[tns] (c) at (3em,7.668em) {};
\draw (c) -- (ab);
\draw (c) -- (a);
\draw (c) -- (b);
\node (pa) at (0,0) {$\apsnodei$};
\node[par] (pc) at (3em,1.732em) {};
\draw (pc) -- (ab);
\path[>=latex,->]  (pc) edge (pa);
\draw (b) to [out=50,in=330] (pc);
\node (labm) at (13em,-2.5em) {$[L\lpr ]$};
\node (tab) at (13em,0.0em) {$\apsnodei$};
\node (ta) at (10em,4.8em) {$\apsnodei$};
\node (tb) at (16em,4.8em) {$\apsnodei$};
\node[tns] (tc) at (13em,2.868em) {};
\draw (tc) -- (tab);
\draw (tc) -- (ta);
\draw (tc) -- (tb);
\node (tabx) at (13em,9.6em) {$\apsnodei$};
\node[par] (tcx) at (13em,6.532em) {};
\path[>=latex,->]  (tcx) edge (tabx);
\draw (tcx) -- (ta);
\draw (tcx) -- (tb);
\node (labr) at (23em,-2.5em) {$[R\ldl ]$};
\node (ab) at (23em,4.8em) {$\apsnodei$};
\node (a) at (26em,9.6em) {$\apsnodei$};
\node (b) at (20em,9.6em) {$\apsnodei$};
\node[tns] (c) at (23em,7.668em) {};
\draw (c) -- (ab);
\draw (c) -- (a);
\draw (c) -- (b);
\node (pa) at (26em,0) {$\apsnodei$};
\node[par] (pc) at (23em,1.732em) {};
\draw (pc) -- (ab);
\path[>=latex,->]  (pc) edge (pa);
\draw (b) to [out=130,in=210] (pc);
\end{tikzpicture}
\end{center}
\caption{Contractions --- Lambek connectives}
\label{fig:lambek_contr}
\end{figure}
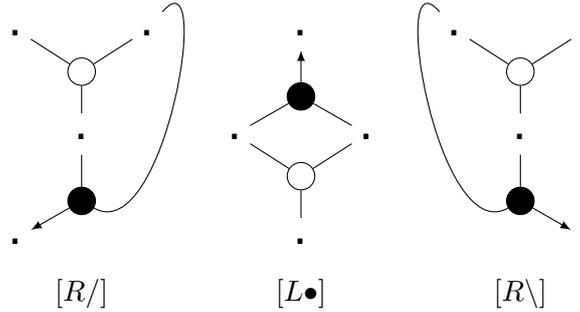

The
contractions for NL are shown in Figure~\ref{fig:lambek_contr}. Each
destructor is drawn in black and has a single outgoing arrow (this
arrow points to the main formula of the link). 

We can contract only 
when the branches of a constructor and destructor are connected by the 
two branches without the arrow. For the $[L\bullet]$ contraction, this
means we connect the two left branches to each other and similarly for
the two right branches. The contractions for the implications look a
bit odd when we draw them on the plane and preserve left/right and
top/bottom of the links. The $[R/]$ contraction connects a constructor
and destructor top-to-bottom while connecting the two right branches
(the left branch has the arrow pointing towards it). Once we have
connected top and bottom, we are forced to bend one of the right
branches (alternatively, we could write the graph on a cylinder).
The idea behind the  $[R/]$ contraction is that it withdraws a hypothesis
occurring as the immediate right daughter of the root of the tree,
just like the sequent calculus or natural deduction rule for NL would
do. 
%
%
The bent link can be seen as a graphical representation of the coindexing of the hypothesis with
the rule application for the $/I$ rule. By simple pattern-matching with
Figure~\ref{fig:nll} we can see that these simple instances of the $[R/]$  contraction
corresponds to proofs of $A\ldr B \vdash A\ldr B$, $A \vdash (A\lpr
B)\ldr B$ and $A \vdash C\ldr (A\ldl C)$.

The important idea of take away from this is that we specify our
structures (in the case of NL, these were simple binary branching trees),
and that from this we automatically obtained three connectives, where each of
the three possible ``main edges'' induced a corresponding
contraction. 

Moving from NL to the multimodal Lambek calculus NL$\Diamond_{\mathcal
  R}$ means changing our structures from unlabelled binary trees to
labelled binary-unary trees, that is labelled trees where both binary
and unary branches are allowed. The unary branches corresponds to the
connectives $\Box$ and $\Diamond$, and the labels, which we draw inside
the central circle of a branch, correspond to mode information. The
contractions have the additional condition that the modes of the
constructor and destructor must be the same.

The final component of the calculus is the set of structural rules,
which are rewrites from one subtree of the graph to another. The unary
branches and the mode information control which structural
rules can apply, so we can have a non-associative base logic yet still
handle right-node-raising and quantifier scope
\cite{KurtoMM,mac10}. Another way to see the structural rules it that
they specify a partial order on our structures. 

For reasons of space, this introduction to the graph rewriting
approach to proof nets is a bit impressionistic, for much more detail and
the fundamental results, the reader is referred to
\cite{mp,mr12lcg}.
Though
created as a proof net calculus for multimodal categorial grammars,
 it can easily be generalised to handle the Lambek-Grishin calculus,
 which adds up-down symmetric structures to NL (see the left hand side
 of Figure~\ref{fig:lg}), and the Lambek calculus
with Galois connections \cite{galois}, which adds negation-like polarity changes to
the calculus (see the middle of Figure~\ref{fig:lg}). Other calculi
also have a presentation which is equivalent to a multimodal calculus,
so we can use this proof net calculus also for \cites{ov14} multimodal
version of the Displacement calculus 
and for $\textrm{NL}_{CL}$, a multimodal version of
$\textrm{NL}_{\lambda}$ \citep[Chapter~17]{bs14cont}, which adds
zero-ary connectives  (``constants'') to the multimodal calculus. All
these proof net calculi are simple instances of the general methodology: specify the
structures (and possibly a partial order on them) and obtain the
connectives and the  contractions for a proof net calculus  \citep{moot07display}.

In the next sections, I illustrate with the help of two examples how this proof net calculus allows us to
compare different systems. 

\subsection{Lambek-Grishin and Galois connections}

\begin{figure}
\begin{center}
\begin{tikzpicture}[node distance=3.75mm]
\node[tns] (c) at (2em,2em) {};
\node (b) at (2em,0em) {$\apsnodei$};
\node (l) at (0em,3.6em) {$\apsnodei$};
\node (r) at (4em,3.6em) {$\apsnodei$};
\draw (b) -- (c);
\draw (c) -- (l);
\draw (c) -- (r);
\node[tns] (c2) at (8em,1.6em) {};
\node (b2) at (8em,3.6em) {$\apsnodei$};
\draw (c2) -- (b2);
\node (l2) at (6em,0em) {$\apsnodei$};
\node (r2) at (10em,0em) {$\apsnodei$};
\draw (l2) -- (c2);
\draw (r2) -- (c2);
\node[tns] (c) at (20em,2em) {};
\node (b) at (20em,0em) {$\apsnodei$};
\node (l) at (18em,3.6em) {$\apsnodei$};
\node (r) at (22em,3.6em) {$\apsnodei$};
\draw (b) -- (c);
\draw (c) -- (l);
\draw (c) -- (r);
\node[tns] (gal) at (16em,5.2em) {};
\node (ll) at (14em,3.6em) {$\apsnodei$};
\draw (l) -- (gal);
\draw (gal) -- (ll);
\node (ub) at (2em,-1.5em) {$\underbrace{\qquad\qquad}_\text{NL}$};
\node (ub2) at (5em,-3.0em) {$\underbrace{\qquad\qquad\qquad\qquad\qquad}_\text{LG}$};
\node (ub3) at (18em,-3.0em) {$\underbrace{\qquad\qquad\qquad\qquad}_\text{NL+Galois}$};
\node (ub4) at (29.9em,-3.0em) {$\underbrace{\qquad\qquad\quad}_\text{LG+top}$};
\node[tns] (c2) at (30em,1.6em) {};
\node (b2) at (30em,3.6em) {$\apsnodei$};
\draw (c2) -- (b2);
\node (l2) at (28em,0em) {$\apsnodei$};
\node (r2) at (32em,0em) {$\apsnodei$};
\draw (l2) -- (c2);
\draw (r2) -- (c2);
\node[tns] (top) at (30em,5.2em) {};
\draw (b2) -- (top);
\end{tikzpicture}
\end{center}
\caption{NL, LG and Galois connections}
\label{fig:lg}
\end{figure}
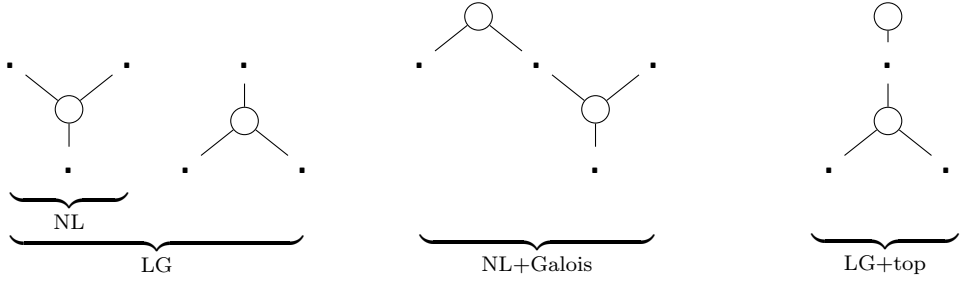

First, we can see that the non-associative
Lambek calculus (NL) extended with Galois connectives has the
Lambek-Grishin calculus as a special case. As Figure~\ref{fig:lg} shows,
LG extends NL with up-down symmetric connectives whereas the Galois
connectives give a negation-like polarity change operation. In
particular, as shown in the center of Figure~\ref{fig:lg}, this allows
us to simulate the Grishin connectives: a combination of an NL
constructor with a Galois connective as shown in the figure forms a
subgraph which to the outside ``looks the same'', up until the cyclic
order of the external vertices, as the Grishin connective  (there is
an alternative solution, which is left-right symmetric to the one
shown). 

This mapping goes only in one direction, however. The
translation back to LG is partial: though the images of translations
of LG graphs are sent back to the original graph, a graph consisting
of a single Galois connective cannot be translated to LG, since only
the ``unit'' displayed in the center of the figure is a meaningful
candidate for such a translation. 

However, suppose there is something interesting in NL+Galois and we
are interested in translating this analysis to LG. The graphs
immediately suggest the solution shown on the right of
Figure~\ref{fig:lg}: a combination of a Grishin connective with a
0-ary ``top'' connective can function as a Galois connective. Like before the
two graphs look the same to the outside world: both have no premisses and two
conclusions in the same order (a 0-ary connective is not the same as
an identity element: if we want such a connective to function as an
identity element for a binary connective we need to add the
appropriate structural rules, an example of such a structural rule can
be found on the left of Figure~\ref{fig:id} below).

To complete the picture, we have the following relations between NL,
LG and the Galois/top connectives ($A\subset B$ means here that we can
translate structures of logic $A$ into structures of logic $B$ in such
a way that logic $B$ is a conservative extension of logic $A$).

\begin{center}
 NL $\subset$ LG $\subset$ NL+Galois $\subset$ LG+top = NL+Galois+top
\end{center}

If we want a fully up-down symmetrical system, we need to complete the
system with  dual-Galois connectives and a bottom connective in the
obvious way.

\subsection{Multimodal versions of $q$}

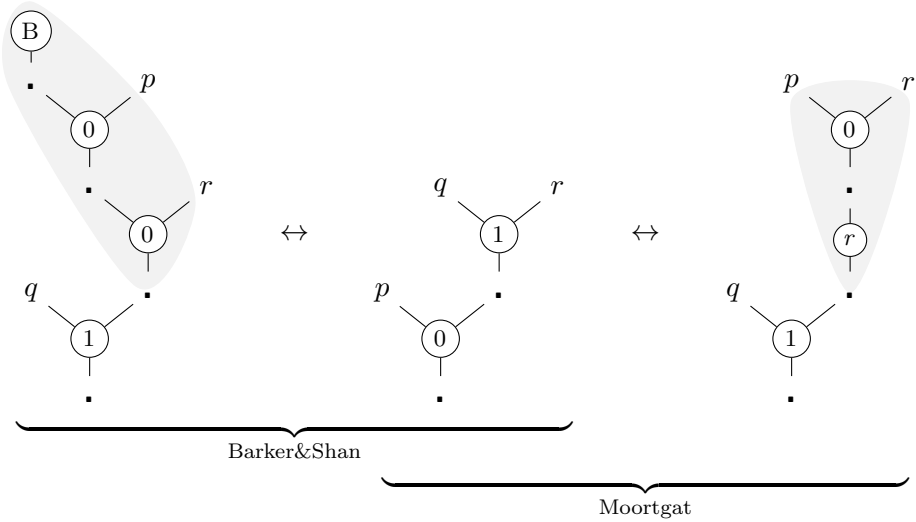
\begin{figure}
\begin{center}
\begin{tikzpicture}[node distance=3mm]
\node[tns] (c) at (2em,2em) {\myindex{0}};
\node (b) at (2em,0em) {$\apsnodei$};
\node (l) at (0em,3.6em) {$p$};
\node (r) at (4em,3.6em) {$\apsnodei$};
\draw (b) -- (c);
\draw (c) -- (l);
\draw (c) -- (r);
\node[tns] (c2) at (4em,5.6em) {\myindex{1}};
\draw (r) -- (c2);
\node (d) at (2em,7.2em) {$q$};
\node (e) at (6em,7.2em) {$r$};
\draw (c2) -- (d);
\draw (c2) -- (e);
\draw [fill = lightgray,color=lightgray] plot [smooth cycle] coordinates { (14em,10.2em) (18em,10.2em) (16em,3.6em) };
\node[tns] (c) at (14em,2em) {\myindex{1}};
\node (b) at (14em,0em) {$\apsnodei$};
\node (l) at (12em,3.6em) {$q$};
\node (r) at (16em,3.6em) {$\apsnodei$};
\draw (b) -- (c);
\draw (c) -- (l);
\draw (c) -- (r);
\node (r2) at (16em,7.2em) {$\apsnodei$};
\node[tns] (c3) at (16em,5.4em) {\myindexi{r}};
\node[tns] (c2) at (16em,9.2em) {\myindex{0}};
\draw (r) -- (c3);
\draw (c2) -- (r2);
\draw (r2) -- (c3);
\node (d) at (14em,10.8em) {$p$};
\node (e) at (18em,10.8em) {$r$};
\draw (c2) -- (d);
\draw (c2) -- (e);
\draw [fill = lightgray,color=lightgray] plot [smooth cycle] coordinates { (-12em,13.6em) (-8em,10.0em) (-6.4em,6.6em)  (-8.4em,3.8em) (-12.8em,10.8em)};
\node[tns] (c) at (-10em,2em) {\myindex{1}};
\node (b) at (-10em,0em) {$\apsnodei$};
\node (l) at (-12em,3.6em) {$q$};
\node (r) at (-8em,3.6em) {$\apsnodei$};
\draw (b) -- (c);
\draw (c) -- (l);
\draw (c) -- (r);
\node (r2) at (-10em,7.2em) {$\apsnodei$};
\node[tns] (c3) at (-8em,5.6em) {\myindex{0}};
\node[tns] (c2) at (-10em,9.2em) {\myindex{0}};
\draw (r) -- (c3);
\draw (c2) -- (r2);
\draw (r2) -- (c3);
\node (d) at (-8em,10.8em) {$p$};
\node (e) at (-6em,7.2em) {$r$};
\draw (c2) -- (d);
\draw (c3) -- (e);
\node (f) at (-12em,10.8em) {$\apsnodei$};
\node[tns] (c4) at (-12em,12.6em) {\myindex{B}};
\draw (c4) -- (f);
\draw (f) -- (c2);
\node (ub) at (-3em,-1.5em) {$\underbrace{\qquad\qquad\qquad\qquad\qquad\qquad\qquad\qquad\qquad\quad}_\text{Barker\&Shan}$};
\node (ub2) at (9em,-3.5em) {$\underbrace{\qquad\qquad\qquad\qquad\qquad\qquad\qquad\qquad\qquad}_\text{Moortgat}$};
\node (lr1) at (-3em,5.6em) {$\leftrightarrow$};
\node (lr2) at (9em,5.6em) {$\leftrightarrow$};
\end{tikzpicture}
\end{center}
\caption{Mixed commutativity in \citep{bs14cont}  compared to
  \citep{mac10}}
\label{fig:mc}
\end{figure}

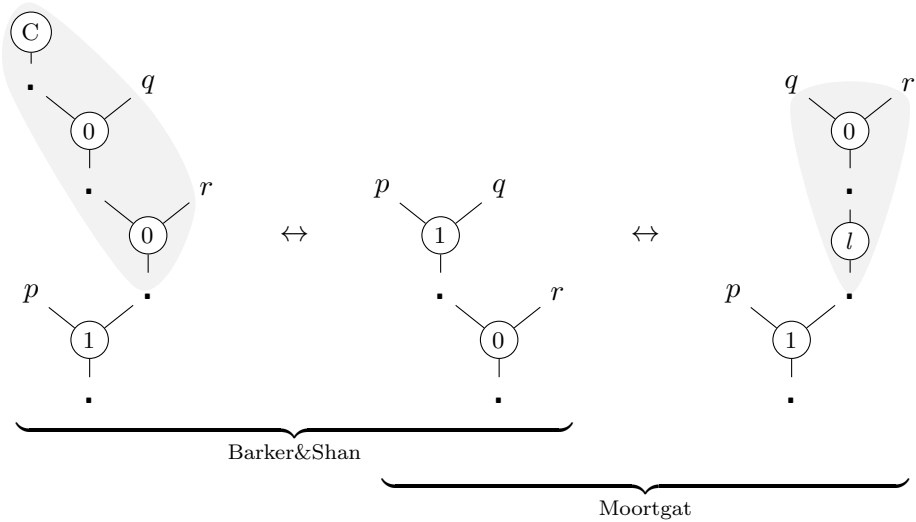
\begin{figure}
\begin{center}
\begin{tikzpicture}[node distance=3mm]
\node[tns] (c) at (4em,2em) {\myindex{0}};
\node (b) at (4em,0em) {$\apsnodei$};
\node (l) at (6em,3.6em) {$r$};
\node (r) at (2em,3.6em) {$\apsnodei$};
\draw (b) -- (c);
\draw (c) -- (l);
\draw (c) -- (r);
\node[tns] (c2) at (2em,5.6em) {\myindex{1}};
\draw (r) -- (c2);
\node (d) at (0em,7.2em) {$p$};
\node (e) at (4em,7.2em) {$q$};
\draw (c2) -- (d);
\draw (c2) -- (e);
\draw [fill = lightgray,color=lightgray] plot [smooth cycle] coordinates { (14em,10.2em) (18em,10.2em) (16em,3.6em) };
\node[tns] (c) at (14em,2em) {\myindex{1}};
\node (b) at (14em,0em) {$\apsnodei$};
\node (l) at (12em,3.6em) {$p$};
\node (r) at (16em,3.6em) {$\apsnodei$};
\draw (b) -- (c);
\draw (c) -- (l);
\draw (c) -- (r);
\node (r2) at (16em,7.2em) {$\apsnodei$};
\node[tns] (c3) at (16em,5.4em) {\myindexi{l}};
\node[tns] (c2) at (16em,9.2em) {\myindex{0}};
\draw (r) -- (c3);
\draw (c2) -- (r2);
\draw (r2) -- (c3);
\node (d) at (14em,10.8em) {$q$};
\node (e) at (18em,10.8em) {$r$};
\draw (c2) -- (d);
\draw (c2) -- (e);
\draw [fill = lightgray,color=lightgray] plot [smooth cycle] coordinates { (-12em,13.6em) (-8em,10.0em) (-6.4em,6.6em)  (-8.4em,3.8em) (-12.8em,10.8em)};
\node[tns] (c) at (-10em,2em) {\myindex{1}};
\node (b) at (-10em,0em) {$\apsnodei$};
\node (l) at (-12em,3.6em) {$p$};
\node (r) at (-8em,3.6em) {$\apsnodei$};
\draw (b) -- (c);
\draw (c) -- (l);
\draw (c) -- (r);
\node (r2) at (-10em,7.2em) {$\apsnodei$};
\node[tns] (c3) at (-8em,5.6em) {\myindex{0}};
\node[tns] (c2) at (-10em,9.2em) {\myindex{0}};
\draw (r) -- (c3);
\draw (c2) -- (r2);
\draw (r2) -- (c3);
\node (d) at (-8em,10.8em) {$q$};
\node (e) at (-6em,7.2em) {$r$};
\draw (c2) -- (d);
\draw (c3) -- (e);
\node (f) at (-12em,10.8em) {$\apsnodei$};
\node[tns] (c4) at (-12em,12.6em) {\myindex{C}};
\draw (c4) -- (f);
\draw (f) -- (c2);
\node (ub) at (-3em,-1.5em) {$\underbrace{\qquad\qquad\qquad\qquad\qquad\qquad\qquad\qquad\qquad\quad}_\text{Barker\&Shan}$};
\node (ub2) at (9em,-3.5em) {$\underbrace{\qquad\qquad\qquad\qquad\qquad\qquad\qquad\qquad\qquad}_\text{Moortgat}$};
\node (lr1) at (-3em,5.6em) {$\leftrightarrow$};
\node (lr2) at (9em,5.6em) {$\leftrightarrow$};
\end{tikzpicture}
\end{center}
\caption{Mixed associativity in \citep{bs14cont}  compared to
  \citep{mac10}}
\label{fig:ma}
\end{figure}

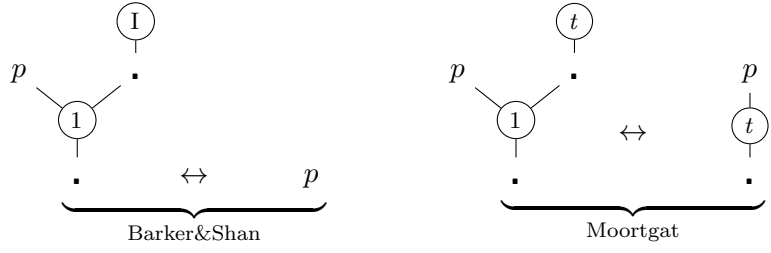
\begin{figure}
\begin{center}
\begin{tikzpicture}[node distance=3mm]
\node[tns] (c) at (1em,2em) {\myindex{1}};
\node (b) at (1em,0em) {$\apsnodei$};
\node (l) at (-1em,3.6em) {$p$};
\node (r) at (3em,3.6em) {$\apsnodei$};
\draw (b) -- (c);
\draw (c) -- (l);
\draw (c) -- (r);
\node[tns] (e) at (3em,5.4em) {\myindex{I}};
\draw (r) -- (e);
\node (p) at (9em,0em) {$p$};
\node[tns] (c) at (16em,2em) {\myindex{1}};
\node (b) at (16em,0em) {$\apsnodei$};
\node (l) at (14em,3.6em) {$p$};
\node (r) at (18em,3.6em) {$\apsnodei$};
\draw (b) -- (c);
\draw (c) -- (l);
\draw (c) -- (r);
\node[tns] (e) at (18em,5.4em) {\myindexi{t}};
\draw (r) -- (e);
\node[tns] (c) at (24em,1.8em) {\myindexi{t}};
\node (b) at (24em,0em) {$\apsnodei$};
\node (l) at (24em,3.6em) {$p$};
\draw (b) -- (c);
\draw (c) -- (l);
\node (ub) at (5em,-1.5em) {$\underbrace{\qquad\qquad\qquad\qquad\quad}_\text{Barker\&Shan}$};
\node (ub2) at (20em,-1.5em) {$\underbrace{\qquad\qquad\qquad\qquad\quad}_\text{Moortgat}$};
\node (lr1) at (5em,0.0em) {$\leftrightarrow$};
\node (lr2) at (20em,1.5em) {$\leftrightarrow$};
\end{tikzpicture}
\end{center}
\caption{Start rules of \citep{bs14cont}  compared to 
  \citep{mac10}}
\label{fig:id}
\end{figure}

As a second illustration, \citeasnoun{mac10} and \citeasnoun{bs14cont}
both provide an implementation of the in situ binder $q(A,B,C)$, shown
in Figure~\ref{fig:mc} and Figure~\ref{fig:ma} on the right and center
resp.\ left and center (both have an additional ``start'' rule shown
in Figure~\ref{fig:id}, discussed below). For the rules of Figures~\ref{fig:mc} and~\ref{fig:ma} the two substructures displayed
in gray give us a mapping from one system to the other. This
highlights both similarities and differences between the two
approaches: the Barker and Shan implementation, which they call $\textrm{NL}_{CL}$, works unchanged when
multiple in situ binders ``overlap'', as used for their analysis of
the semantics of ``\emph{same}'', whereas Moortgat's approach works
correctly even in the presence of associativity. Both properties are a
consequence of the unary branches and their position in Moortgat's
calculus: the unary branches block associativity, but also prevent the
rules from operating in the direction from center to right when there
is a unary branch between the two constructors.

In Moortgat's original system, the
unary branches leave a ``trail'' telling us at each binary branch
whether to go left or right. In the Barker and Shan system, there is a
similar trail
but stored at a deeper position. Barker and Shan's system
breaks down when we add associativity to the system --- though the
addition of modally controlled associativity seems delicate even in
Moortgat's system and needs to be carefully investigated.

Finally, there are the start rules of Figure~\ref{fig:id}. For the
Barker and Shan system the $I$ connective is simply an right identity
element for the binary mode 1 (the structural rule looks strange
compared to the ones we've seen before, but it is of the same shape:
we replace a subtree with leaf $p$, binary mode 1 and 0-ary mode $I$
by the leaf $p$, which is a subtree with the same leaves). In
Moortgat's version, we replace a unary branch $t$ by a binary branch
with a right identity element $t$ ($t$ occurs both as a unary branch
and as a 0-ary branch in this rule). This formulation has the
advantage that we can lexically control the application of the
rule (the current form of these rules is at least partially motivated
from the point of view of facilitating automated deduction for the system). For the Barker and Shan implementation, we need to be careful
when applying the rule from right-to-left, since naive application of
this rule can lead us into an infinite loop.

To intertranslate the
start rules, we can either use the fact that we can eliminate
``useless'' applications of the $I$ rule from $\textrm{NL}_{CL}$ --- we
know from the embedding results of
Sections~17.5 and~17.7 and Definition~304 on page~172 of
\citep{bs14cont} that we can, without loss of generality, restrict
$\textrm{NL}_{CL}$ structures containing $I$ only at the end of a path
of $B$ and $C$ nodes --- or import Moortgat's analysis directly into
$\textrm{NL}_{CL}$. In either case, it is then a simple translation
between 0-ary $I$ and 0-ary $t$.

Looking at Figures~\ref{fig:mc} and~\ref{fig:ma} also shows us exactly
when we can translate $\textrm{NL}_{CL}$ structures with $BCI$
connectives to Moortgat's calculus: only when $B$ and $C$ nodes occur
as the right daughter of two consecutive $\circ_0$ branches, that is
only when they occur as $(B \circ_0 X)\circ_0 Y$ or $(C\circ_0
X)\circ_0 Y$ (for some structures $X$ and $Y$). But this
property crucially fails when there are multiple binders, such
as in the term below, which occurs in the analysis of ``the same
waiter served everyone'' on page~166 of \citep{bs14cont}.
\[
\textit{same} \circ_1 (( C \circ_0 (( B \circ_0 \mathbf{B}) \circ_0 ((B \circ_0
\textit{the}) \circ_0 ((C\circ_0 I) \circ_0 \textit{waiter}))))
\circ_0 ( (B\circ_0 \textit{served}) \circ_ 0 I))
\]

In the above tree (presented as a flat term to preserve space) the $B$
indicated in bold is not translated when we simply intertranslate the structures
indicated in gray in Figures~\ref{fig:mc} and~\ref{fig:ma}, since it
occurs on a right branch. The translation produces the following term
which still contains this $B$ which is not a part of Moortgat's
calculus (the circumfix $\langle . \rangle^i$ denotes a unary branch with
label $i$).
\[
\textit{same} \circ_1 \langle \langle \mathbf{B} \circ_0 \langle \textit{the} \circ_0
\langle t\circ_0 \textit{waiter} \rangle^l \rangle^r \rangle^r \circ_0 \langle \textit{served} \circ_0 t \rangle^r \rangle^l
\]

In the Barker and Shan analysis, this $B$ is part of the second path; after
infixation of ``same'' using the structural rules, we obtain the following
term, with the same $B$ shown in bold.
\[
(\mathbf{B} \circ_0 (\textit{the} \circ_0 (\textit{same} \circ_0 \textit{waiter})))
\circ_0 ( (B\circ_0 \textit{served}) \circ_ 0 I)
\]

This second term translates unproblematically to the following term.
\[\langle
(\textit{the} \circ_0 (\textit{same} \circ_0 \textit{waiter})) \circ_0
\langle 
\textit{served} \circ_0 t \rangle^r \rangle^r
\]

Though this analysis does not immediately show us the best way to extend
Moortgat's calculus to handle multiple binders, it does give us an
indication of where to look: we want to extend the translation in such
a way that $B$ terms can no longer appear and which allows us to
rewrite our new terms, no longer containing $B$, to the term above. It is not hard to add modes
and structural rules to allow exactly this, but I'll leave more
elegant solutions to future research.

\subsection{Graph rewriting, graph grammars and modal logic}

I have briefly talked about how graph rewriting can be a tool for
providing mappings from one type-logical grammar to another. There are
some other connections I briefly want to mention. 

Since most type-logical grammars have models in modal logic, can we
use some of the rich set of tools of modal logic to discover relations
between different formalisms? For example by using bisimulations
between the model-theoretic interpretations of two calculi
\cite{ML01}. This would be a model-theoretic counterpart to the
proof-theoretic approach of the current paper. Also, since the modern
view of modal logic sees modal logic as (decidable) fragments of
first-order logic this ties in nicely with the first-order perspective
of the next section\footnote{I thank one of the anonymous referees for
  pointing out the connection to modal logic.}.

A second point is that graph rewriting has been extensively studied in
computer science. However, these tend to be nicer classes of graphs
than those presented here. Is there a connection between some
context-free graph grammars \citep{engelfriet97cfgg} and the proof nets
studied here or do we, as seems likely, need richer classes? If so,
which ones and what are their properties? Some preliminary investigations into this topic can be
found in \citep{moot08hyper}.

\section{First-order linear logic}

The second meta-logic is first-order (multiplicative, intuitionistic)
linear logic. Though the idea of adding first-order quantifiers to
type-logical grammars is very old \citep[Section 6.2]{morrill}, the important shift in \citeasnoun{mill1} was that the first-order quantifiers could also be used for
word order. So instead of adding first-order quantifiers to one of the
other variants of type-logical grammar, we can add first-order
quantifiers to multiplicative linear logic and embed the Lambek
calculus. Several of the known problems of
the Lambek calculus have a simple solution in first-order linear
logic. First-order linear logic also has a very nice, clean proof net
calculus \cite{quant}.

\subsection{Proof search in first-order linear logic}

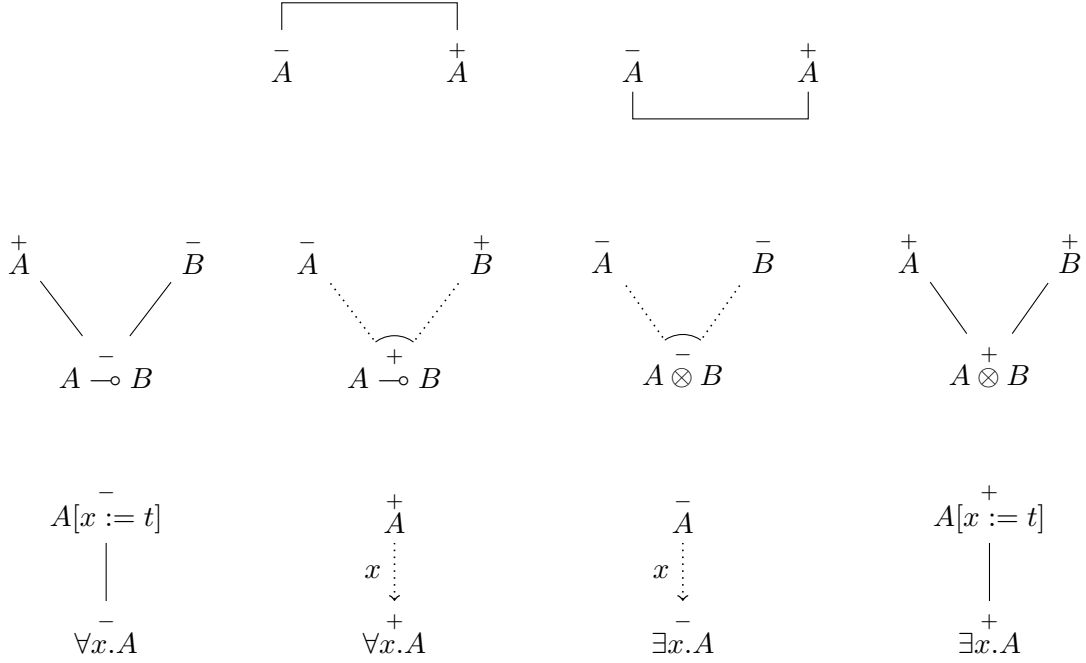
\begin{figure}
\begin{center}
\begin{tikzpicture}
\node (anx) at (0em,0em) {$\overset{-}{A}$};
\node (anp) at (6em,0em) {$\overset{+}{A}$};
\draw (anx) -- (0em,2em) -- (6em,2em) -- (anp);
\node (cnx) at (12em,0em) {$\overset{-}{A}$};
\node (cnp) at (18em,0em) {$\overset{+}{A}$};
\draw (cnx) -- (12em,-2em) -- (18em,-2em) -- (cnp);
\end{tikzpicture}
\end{center}
\vspace{\baselineskip}
\begin{center}
\begin{tikzpicture}
\node (forallnc) {$\overset{-}{\forall x. A}$};
\node (forallnp) [above=2em of forallnc] {$\overset{-}{A[x:=t]}$};
\draw (forallnc) -- (forallnp);
\node (forallpc) [right=7em of forallnc] {$\overset{+}{\forall x. A}$};
\node (forallpp) [above=2em of forallpc] {$\overset{+}{A}$};
\draw [<-,semithick,dotted] (forallpc) -- (forallpp) node [midway,left=1pt] {$x$};
\node (existsnc) [right=7em of forallpc] {$\overset{-}{\exists x. A}$};
\node (existsnp) [above=2em of existsnc] {$\overset{-}{A}$};
\draw [<-,semithick,dotted] (existsnc) -- (existsnp)  node [midway,left=1pt] {$x$};
\node (otimesnc) [above=7em of existsnc] {$\overset{-}{A\otimes B}$};
\node (tmponl) [left=0.66em of otimesnc] {};
\node (aotimesnc) [above=2.5em of tmponl] {$\overset{-}{A}$};
\node (tmponr) [right=0.66em of otimesnc] {};
\node (botimesnc) [above=2.5em of tmponr] {$\overset{-}{B}$};
\begin{scope}
\begin{pgfinterruptboundingbox}
\path [clip] (otimesnc.center) circle (2.5ex) [reverseclip];
\end{pgfinterruptboundingbox}
\draw [semithick,dotted] (otimesnc.center) -- (botimesnc);
\draw [semithick,dotted] (otimesnc.center) -- (aotimesnc);
\end{scope}
\begin{scope}
\path [clip] (aotimesnc) -- (otimesnc.center) -- (botimesnc);
\draw (otimesnc.center) circle (2.5ex);
\end{scope}
\node (otimespc) [right=7em of otimesnc] {$\overset{+}{A\otimes B}$};
\node (tmpopl) [left=0.66em of otimespc] {};
\node (aotimespc) [above=2.5em of tmpopl] {$\overset{+}{A}$};
\node (tmpopr) [right=0.66em of otimespc] {};
\node (botimespc) [above=2.5em of tmpopr] {$\overset{+}{B}$};
\draw (otimespc) -- (aotimespc);
\draw (otimespc) -- (botimespc);
\node (existspc) [below=7em of otimespc] {$\overset{+}{\exists x. A}$};
\node (existspp) [above=2em of existspc] {$\overset{+}{A[x:=t]}$};
\draw  (existspc) -- (existspp);
\node (lollinc) [above=7em of forallnc] {$\overset{-}{A\multimap B}$};
\node (tmplnl) [left=0.66em of lollinc] {};
\node (alollin) [above=2.5em of tmplnl] {$\overset{+}{A}$};
\draw (lollinc) -- (alollin);
\node (tmplnr) [right=0.66em of lollinc] {};
\node (blollin) [above=2.5em of tmplnr] {$\overset{-}{B}$};
\draw (lollinc) -- (blollin);
\node (lollipc) [above=7em of forallpc] {$\overset{+}{A\multimap B}$};
\node (tmplpl) [left=0.66em of lollipc] {};
\node (alollip) [above=2.5em of tmplpl] {$\overset{-}{A}$};
\node (tmplpr) [right=0.66em of lollipc] {};
\node (blollip) [above=2.5em of tmplpr] {$\overset{+}{B}$};
\begin{scope}
\begin{pgfinterruptboundingbox}
\path [clip] (lollipc.center) circle (2.5ex) [reverseclip];
\end{pgfinterruptboundingbox}
\draw [semithick,dotted] (lollipc.center) -- (blollip);
\draw [semithick,dotted] (lollipc.center) -- (alollip);
\end{scope}
\begin{scope}
\path [clip] (alollip) -- (lollipc.center) -- (blollip);
\draw (lollipc.center) circle (2.5ex);
\end{scope}
\end{tikzpicture}
\end{center}

\caption{Logical links for MILL1 proof structures}
\label{tab:links}
\end{figure}

I will give a very brief introduction to first-order linear logic from
the point of view of proof search and parsing.
Figure~\ref{tab:links} presents the logical links for first-order
linear logic proof nets. To try and prove a sequent $A_1,\ldots,A_n
\vdash C$ we decompose the formulas according to the links of the
figure, starting with 

$$
\overset{-}{A_1} \ldots \overset{-}{A_n} \,\overset{+}{C}
$$

\noindent and continuing the unfolding until we reach the atomic
subformulas. The links essentially produce a subformula tree with the
additional marking of the polarity of each subformula and with some
dotted lines. There are two types of dotted lines: those which pair two branches (for the positive $\multimap$ and
negative $\otimes$ link) and those which are labelled with the eigenvariable
of a quantifier (for the positive $\forall$ and the negative $\exists$
link). We use the standard convention that each quantifier in a
sequent has a different eigenvariable.

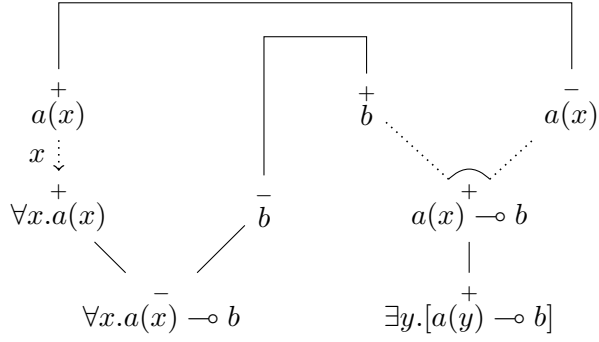
\begin{figure}
\begin{center}
\begin{tikzpicture}[scale=0.9]
\node (b1) at (1.5cm,0cm) {$\overset{-}{\forall x. a(x)  \multimap b}$};
\node (lax)  at (0cm,1.5cm) {$\overset{+}{\forall x. a(x)}$};
\node (lb) at (3cm,1.5cm)  {$\overset{-}{b}$};
\node (g) at (6cm,0cm) {$\overset{+}{\exists y. [ a(y) \multimap b ] }$};
\node (g2) at (6cm,1.5cm) {$\overset{+}{a(x) \multimap b}$};
\draw (g) -- (g2);
\node  (ga) at (7.5cm,3cm) {$\overset{-}{a(x)}$};
\node  (gb) at (4.5cm,3cm) {$\overset{+}{b}$};
\node (la) at (0cm,3cm) {$\overset{+}{a(x)}$};
\draw (ga) -- (7.5cm,4.5cm) -- (0cm,4.5cm) -- (la);
\draw (gb) -- (4.5cm,4cm) -- (3cm,4cm) -- (lb);
\draw[<-,semithick,dotted] (lax) -- (la)  node [midway,left=1pt] {$x$};
\draw (b1) -- (lax);
\draw (b1) -- (lb);
\draw[semithick,dotted] (6.3cm,1.9cm) -- (ga);
\draw[semithick,dotted] (5.7cm,1.9cm) -- (gb);
\draw plot [smooth,tension=1] coordinates {(5.7cm,1.9cm) (6cm,2.05cm) (6.3cm,1.9cm)};
\end{tikzpicture}
\end{center}
\caption{Proof structure which is not a proof net}
\label{fig:noproof}
\end{figure}

After this unfolding step, we connect atomic formulas of opposite
polarity using the axiom link (Figure~\ref{tab:links}, top left). When
all atomic formulas have been connected this way, the resulting
structure is called a proof structure. As
an example, the sequent $\forall x. a(x)  \multimap b \vdash  \exists
y. [ a(y) \multimap b ]$ has the proof structure shown in Figure~\ref{fig:noproof}.

In practice, we do not choose a term for the negative $\forall$ or for
the positive $\exists$ link during formula unfolding but we rather use
meta-variables for unfolding and unification during the axiom link
connections. So the axiom link does not connect identical formulas but
rather unifiable formulas and performs this unification. This is a
rather standard theorem-proving strategy and has the result that we
can read off the most general term for each negative $\forall$ and
positive $\exists$ rule in our proof net. Some care must be taken when
we repeatedly unify the positive and negative atomic subformulas of $\forall x. a(x) \multimap
a(f(x,x))$, where the size of the term argument grows exponentially in
the number of occurrences of the given formula (
$a(x)$, $a(f(x,x))$, $a(f(f(x,x),f(x,x)))$,
$a(f(f(f(x,x),f(x,x)),f(f(x,x),f(x,x))))$, $\ldots$) . Even in these cases, we can ensure linear time unification
by adopting a sharing strategy
\citep{martelli1982efficient,pw78unif}. In addition, for the typical uses
of first-order linear logic which interest us here, each quantifier
binds at most two occurrences of its eigenvariable
\cite{moot13lambek}, so we might even decide to exclude the case above, where the quantifier $\forall
x$ binds three occurrences.

Returning to the example proof structure of Figure~\ref{fig:noproof},
the given sequent is underivable in linear logic and, in general, proof
structures need not correspond to proofs. Proof
structures which correspond to proofs are called \emph{proof nets} and we can
distinguish them from other proof structures using properties of the
graph. \citeasnoun{quant} gives a switching criterion whereas
\citeasnoun{moot13lambek} presents a contraction criterion for
first-order linear logic in the style of \citeasnoun{reductions}.

\begin{figure}
\begin{center}
\begin{tikzpicture}
\node (x) at (0em,0em) {$v_i$};
\node (y) at (0em,4em) {$v_j$};
\draw [semithick,dotted] plot [smooth, tension=1] coordinates {(-0.4em,0.5em) (-1em,2em) (-0.4em,3.5em)};
\draw [semithick,dotted] plot [smooth, tension=1] coordinates {(0.4em,0.5em) (1em,2em) (0.4em,3.5em)};
\draw plot [smooth,tension=1] coordinates {(-0.4em,0.5em) (0em,0.7em) (0.4em,0.5em)};
%
\node (x2) at (4em,2em) {$v_i$};
\node (a1) at (2.5em,2em) {$\Rightarrow_{\textit{p}}$};
\node (x3) at (10em,0em) {$v_i$};
\node (y3) at (10em,4em) {$v_j$};
\draw[<-,semithick,dotted] (x3) -- (y3);
\node (xv) at (9.45em,2.15em) {$x$};
\node (x4) at (14em,2em) {$v_i$};
\node (a2) at (12em,2em) {$\Rightarrow_{\textit{u}}$};
\node (x3) at (20em,0em) {$v_i$};
\node (y3) at (20em,4em) {$v_j$};
\draw (x3) -- (y3);
\node (x4) at (24em,2em) {$v_i$};
\node (a3) at (22em,2em) {$\Rightarrow_{\textit{c}}$};
\end{tikzpicture}
\end{center}
\caption{Contractions for first-order linear logic. Conditions: $v_i \neq v_j$
  and, for the $u$ contraction, all free occurrences of $x$ are at $v_j$.}
\label{fig:controne}
\end{figure}
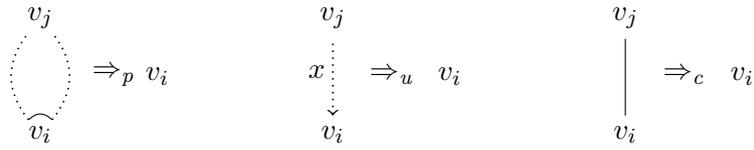

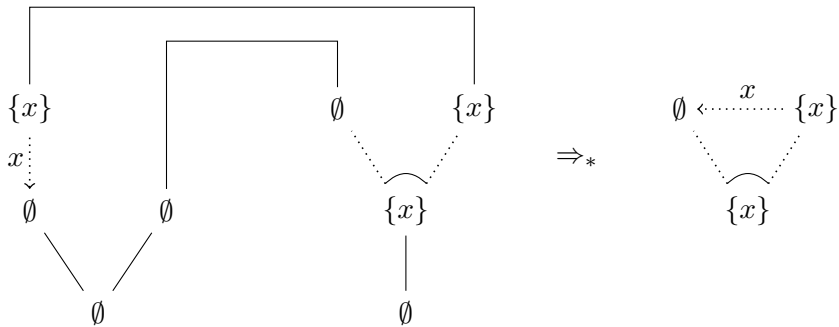
\begin{figure}
\begin{center}
\begin{tikzpicture}[scale=0.9]
\node (b1) at (1.5cm,0cm) {$\emptyset$};
\node (lax)  at (0.5cm,1.5cm) {$\emptyset$};
\node (lb) at (2.5cm,1.5cm)  {$\emptyset$};
\node (g) at (6cm,0cm) {$\emptyset$};
\node (g2) at (6cm,1.5cm) {$\{ x \}$};
\draw (g) -- (g2);
\node  (ga) at (7.0cm,3cm) {$\{ x\}$};
\node  (gb) at (5.0cm,3cm) {$\emptyset$};
\node (la) at (0.5cm,3cm) {$\{ x\}$};
\draw (ga) -- (7.0cm,4.5cm) -- (0.5cm,4.5cm) -- (la);
\draw (gb) -- (5.0cm,4cm) -- (2.5cm,4cm) -- (lb);
\draw[<-,semithick,dotted] (lax) -- (la);
\draw (b1) -- (lax);
\draw (b1) -- (lb);
\draw[semithick,dotted] (6.3cm,1.9cm) -- (ga);
\draw[semithick,dotted] (5.7cm,1.9cm) -- (gb);
\draw plot [smooth,tension=1] coordinates {(5.7cm,1.9cm) (6cm,2.05cm)
  (6.3cm,1.9cm)};
\node (lx) at (0.3cm,2.25cm) {$x$};
\node (arrow) at (8.5cm,2.25cm) {$\Rightarrow_*$};
\node  (ga) at (12.0cm,3cm) {$\{ x\}$};
\node  (gb) at (10.0cm,3cm) {$\emptyset$};
\node (g2) at (11cm,1.5cm) {$\{ x \}$};
\draw[->,semithick,dotted] (ga) -- (gb);
\draw[semithick,dotted] (11.3cm,1.9cm) -- (ga);
\draw[semithick,dotted] (10.7cm,1.9cm) -- (gb);
\draw plot [smooth,tension=1] coordinates {(10.7cm,1.9cm) (11cm,2.05cm)
  (11.3cm,1.9cm)};
\node (lx) at (11.0cm,3.25cm) {$x$};
\end{tikzpicture}
\end{center}
\caption{Contractions for the proof structure of Figure~\ref{fig:noproof}}
\label{fig:noproofb}
\end{figure}

The contractions for first-order linear logic are shown in
Figure~\ref{fig:controne}. As a first step, we forget about all
formulas in the proof structure keeping track only of the free
variables at each vertex in the graph. For Figure~\ref{fig:noproof}
this produces the graph shown on the left hand side of Figure~\ref{fig:noproofb}.
Then we progressively shrink the proof structure using
the contractions shown in the figure. Each contraction removes an edge (a linked pair
of edges in the case of the $p$ contraction) and identifies the two
nodes which were connected by it. A contraction can only be performed
on two distinct vertices, that is, we are not allowed to eliminate
self-loops. The free variables of the result vertex are the union of
the sets of free variables of the two input vertices (in the case of
the $u$ contraction, we can remove the $x$ variable, since it has become redundant). A proof
structure is a proof net if and only if it contracts to a single point
using the contractions of Figure~\ref{fig:controne}.

As an example, Figure~\ref{fig:noproofb} shows the contractions
performed on the proof structure of Figure~\ref{fig:noproof}, with the
initial structure on the left and the structure after all $c$
contractions on the right. The
arrow and eigenvariable of the $\forall$ link and the connection
between the two other dotted links ensure the notation is
unambiguous. The displayed graph is not a single vertex but it cannot be further contracted: the
universal contraction $u$ cannot apply since the variable $x$ occurs
at the bottom vertex instead of only at the right vertex as required
for the contraction and the contraction $p$ cannot apply
until its two branches end at the same vertex. Since the contractions
of Figure~\ref{fig:controne} are confluent\footnote{As discussed in
  \cite{diss}, this is not true for the contractions of
  Section~\ref{sec:graph} which require us to explore the entire
  search space to show underivability. The culprits in that case are
  the unary contractions and the structural rules.}, any graph which
is not further contractible but is not a unique vertex, such as the
one shown at the right of Figure~\ref{fig:noproofb},
suffices to show that the given sequent is underivable.

Summarising, parsing/proof search in first-order linear logic operates as
follows. 

\begin{enumerate}
\item\label{lex} For each word in the sentence, we find a first-order formula in
  the lexicon. 
\item We unfold a sequent using the rules of Figure~\ref{tab:links}.
\item\label{conn} We connect atomic formulas of opposite polarity, unifying variables.
\item We contract the resulting proof structure to a single vertex.
\end{enumerate}

Combinatorially, the complex steps are step~\ref{lex} (in the case of
high lexical ambiguity) and step~\ref{conn} where we connect the
atomic formulas. For an actual implementation, such as
\cite{moot15l1}, it is therefore desirable to contract early ---
thereby keeping a compact representation of the current state of the
proof --- and develop ways of detecting ``doomed'' configurations,
that is graphs which can never be contracted to a single
vertex, no matter how we continue the construction of our proof structure. Examples of such configurations are connections between a node
and its ancestor with a path of dotted links (this corresponds to a cycle
in the proof structure and though we can validly reduce the size of
this cycle, such a configuration will, at best, end up producing a self-loop) or isolated vertices (an isolated vertex is a
vertex which is not connected to the rest of the graph but which also
doesn't have any unlinked atomic formulas; unless it is the last
vertex of the graph, such a vertex corresponds to a disconnected proof
structure). Combining early failure with a smart backtracking strategy
for selecting which atomic formulas to unify \cite{knuth00dancing}
produces an effective algorithm, though I suspect many improvements
are still possible.

\subsection{First-order linear logic as a grammar formalism}

Whereas \citeasnoun{mill1} show that the Lambek calculus has a natural
translation into first-order linear logic,
in \cite{moot13lambek,hal-00996724}, this idea is further developed and translations are given which show
that Displacement grammars, Hybrid Type-Logical Grammars and lambda
grammars are all natural fragments of first-order linear logic,
providing much simpler proof-theoretic foundations for these
calculi. In addition, the analyses proposed in these different
frameworks agree to a large extent upon translation into first-order
linear logic. The basic idea of the translation into first-order
linear logic is that formulas are assigned pairs of string
positions. For example, an atomic Lambek calculus formula $np$ becomes
an atomic first-order linear logic formula $np(0,1)$ when it spans the
string from position 0 to 1. Similarly, the Lambek calculus formula
$np\ldl s$ spanning the string from 1 to 2 becomes the formula
$\forall x. np(x,1) \multimap s(x,2)$, indicating it is looking for a
noun phrase to its right to form a sentence. Instantiating $x$ to 0,
we can combine it with $np(0,1)$ to derive $s(0,2)$, a sentence spanning string
positions 0 to 2. The translations for Hybrid Type-Logical Grammars
and the Displacement calculus, though based on essentially the same idea, are a bit more involved and will not be
repeated here.

Compared to the graph rewriting of Section~\ref{sec:graph}, this setup
has the advantage that instead of comparing \emph{structures} and
giving mappings from structures to structures --- which has the
difficulty that we need to show these structures behave the same in
\emph{all} contexts --- we can translate to first-order linear logic
and compare \emph{formulas}. This is much easier and more immediate:
we translate to first-order linear logic and compare the formulas we
obtain. 
So we can see, for example, that the treatment of gapping in
Hybrid Type-Logical Grammars and the one of the Displacement calculus,
in spite of being formulated using very different logical primitives,
are actually \emph{equivalent} upon translation into first-order
linear logic (in the sense that the translated formulas are
interderivable). There is no need to specify a translation of the
primitives of one calculus into another nor to provide a mapping from
proofs to proofs.

As a simple example, the lambda grammar/ACG lexical entry $np\multimap s$ with
prosodic term $\lambda S. (S + \textit{sleeps})$ becomes, according to
the translation into
first-order linear logic, the formula $\forall x. np(x,1) \multimap
s(x,2)$, just like the Lambek calculus (and Displacement calculus)
formula $np\ldl s$. Even though these translations follow rather
different paths, they end up at the same destination and it is this agreement on many of the ``basic''
lexical entries which
forms the basis of the comparison of formalisms using first-order
linear logic.

\subsection{Relative pronouns}

As a more interesting example of this way of comparing formulas, here are five
different first-order linear logic formulas expressing extraction.
These formulas would be assigned to a relativiser such as ``which''
occurring at position 3-4.
\begin{align}
\!\!\forall x_0. [ (\forall x_1. [ np(x_1,x_1)] \multimap 
s(4,x_0)) \multimap \forall x_2. [n(x_2,3) \multimap n(x_2,x_0)]] \label{new}\\
\!\!\forall x_0. [ \exists x_1. [ np(x_1,x_1) \multimap 
s(4,x_0)] \multimap \forall x_2. [n(x_2,3) \multimap n(x_2,x_0)]]
& \;\; \textrm{D} \label{f:d} \\
\!\!\forall x_0 \forall x_1 \forall x_2. [ (( np(x_1,x_1) \multimap 
s(4,x_0)) \multimap ( n(x_2,3) \multimap n(x_2,x_0))] &
                                                            \;\; \lambda\,\textrm{-grammar}\label{f:lambda} \\
\!\!\forall x_0. [ \forall x_1. [ np(x_1,4) \multimap 
s(x_1,x_0)] \multimap \forall x_2. [n(x_2,3) \multimap n(x_2,x_0)]]
& \;\; \textrm{L}:(n\backslash n)/ (np\backslash s)\label{l:subj} \\
\!\!\forall x_0. [ \forall x_1. [ np(x_0,x_1) \multimap 
s(4,x_1)] \multimap \forall x_2. [n(x_2,3) \multimap n(x_2,x_0)]]
& \;\;  \textrm{L}:(n\backslash n)/  (s/ np) \label{l:obj}
\end{align}

The first three formulas, though with slightly different scopes for
the quantifiers, intuitively mean that a relative pronoun spanning positions 3-4 is
looking to its left for a noun $n$ (a noun spanning positions $x_2$-3
for some $x_2$ of our choice) and to its right for a sentence $s$,
which itself is missing a noun phrase anywhere (where this sentence
spans positions 4-$x_0$ for some $x_0$ of our choice, the relation
between the position of the $np$ and this sentence is not specified,
though the proof theory will ensure this $np$ will occur ``inside''
the sentence). The result will be a noun from position $x_2$, the
start of the $n$ argument, to $x_0$, the end of the $s$ argument.

The first formula is a possibility which I have not seen
before. Formula~\ref{f:d} is the formula from \cite{mill1} as well as
the translation into first-order linear logic of the extraction
formula for the Displacement calculus (D). Formula~\ref{f:lambda} is
the translation of the lambda grammar lexical entry proposed by
\citeasnoun{muskens01lfg}. Finally, formulas~\ref{l:subj}
and~\ref{l:obj} are the translations of the two Lambek calculus
formulas for peripheral extraction. These formulas are related as
follows (where a directed path between the two formulas denotes
derivability of the target from the source).

\begin{center}
\begin{tikzpicture}
\node (a) {\ref{new}};
\node [right=2em of a] (b) {\ref{f:d}};
\node [right=2em of b] (c) {\ref{f:lambda}};
\node [above right=1.5em of c] (d) {\ref{l:subj}};
\node [below right=1.5em of c] (e) {\ref{l:obj}};
\draw[<-] (e) -- (c); 
\draw[<-] (d) -- (c); 
\draw[->] (a) -- (b); 
\draw[->] (b.15) -- (c.165); 
\draw[<-] (b.345) -- (c.195); 
\end{tikzpicture}
\end{center}

So the formulas of the Displacement calculus and the one proposed
directly for first-order linear logic are identical
(formula~\ref{f:d}). This formula is equivalent to formula~\ref{f:lambda}
proposed for $\lambda$-grammars; though we cannot always transform a
linear logic formula into an equivalent prenex
normal form \citep{foll}, formula~\ref{f:d} does allow such a form which is formula~\ref{f:lambda}. When we look at lambda grammars in
isolation, we cannot even directly ask the question about the relation
to Lambek calculus formulas, though here it is clear that formulas~\ref{new}
to~\ref{f:lambda} all have the Lambek calculus formulas~\ref{l:subj}
and~\ref{l:obj} as special cases. The new formula~\ref{new} is the
most general formula, but it is unclear whether or not there is any
useful (or harmful!) difference in behaviour between this formula and
the formulas corresponding to those use in the Displacement calculus
and lambda grammars.

This brings up an important question: since, in (classical) first-order logic,
formula~\ref{new} is equivalent to formulas~\ref{f:d}
and~\ref{f:lambda} maybe first-order \emph{linear} logic is too
fine-grained a tool and the suitable notions of equivalence are better
formulated directly in first-order logic. Is the difference between
classical first-order equivalence and linear first-order equivalence
important, and if, so which is the more suitable notion in the current
context?

\subsection{Adverbs, higher-order formulas and lambda grammars}

The first-order linear logic perspective also clarifies the limitations of abstract
categorial grammars/lambda grammars. For adverbs, for example, we are
looking for a lexical entry which functions at least as well as the
Lambek calculus formula $(np\ldl s)\ldr (np\ldl s)$. However, as shown
in \cite{hal-00996724}, we can simply enumerate all possible ACG lexical
entries $l$, compute their translation into first-order linear logic,
compute the translation of $(np\ldl s)\ldr (np\ldl s)$ into
first-order linear
logic and compare. Keeping only the plausible lexical entries 
(that is, those which generate the right semantics and right word
order) leaves us with three possibilities, which are shown as
items~\ref{delibone} to~\ref{delibthree} below,
together with the translation of $(np\ldl s)\ldr (np\ldl s)$ as
item~\ref{deliblambek} (note the narrow scope of $\forall x_1$ in this
translation). The adverb is assumed to span positions 1-2.
\begin{align}
\label{deliblambek}
\forall x_0 \forall x_2. [ \forall 
  x_1.  [ np(x_1,2) \multimap s(x_1,x_2)] &\multimap (np(x_0,1)
                                            \multimap s(x_0,x_2))] \\
\label{delibone} \forall x_0 \forall x_1 \forall
                                       x_2.  [( np(x_1,x_1) \multimap s(2,x_2)) &\multimap
                                       (np(x_0,1) \multimap s(x_0,x_2))]  \\
\label{delibtwo}  \forall x_0 \forall x_1 \forall
                                       x_2. [ ( np(1,2) \multimap
  s(x_1,x_2)) &\multimap (np(x_0,x_1) \multimap s(x_0,x_2))] \\
\label{delibthree} \forall x_0 \forall x_1 \forall
                                       x_2. [ (np(x_1,2) \multimap
  s(x_0,x_2)) &\multimap (np(x_1,1) \multimap s(x_0,x_2))]
\end{align}

The translations of ACG lexical entries are always formulas with only
universal quantifiers and in prenex normal form\footnote{The term
  Skolem normal form is often used for a prenex normal form with
  universal quantifiers, but I don't use it here since it suggests that
  existential quantifiers have been replaced by universally quantified
  Skolem terms and 1) there are no
  terms in the translations of ACG formulas 2) Skolemization is
  unsound in first-order linear logic. \cite{foll}}.
It is easy to verify that all of items~\ref{delibone}
to~\ref{delibthree} are strictly more general than the translation of $(np\ldl s)\ldr (np\ldl 
s)$, shown as item~\ref{deliblambek}. 

However, where in the case of relative pronouns, a more general 
formula turned out to be a benefit, in the case of adverbs, it turns 
out to be a source of overgeneration. For example,
item~\ref{delibone}, the adverb lexical 
entry most commonly used in the ACG literature, predicts that an adverb 
selects to its right, a sentence missing a noun phrase 
\emph{anywhere}. In other words, the lexical entry for adverbs is 
modelled after the lexical entry for relative pronouns and therefore 
follows a ``medial extraction'' analysis, whereas items~\ref{delibtwo}
and~\ref{delibthree} predict a type of quantifying-in behaviour:
item~\ref{delibtwo} is modelled after the type assigned to a
generalised quantifier but with an extra $np$ argument; it takes as
its argument a sentences missing a noun phrase at the position of the
adverb (just like a generalised quantifier takes a sentence missing an
$np$ at the position of the quantifier as its
argument), making the odd prediction that adverbs occur at the same
place as noun phrases. Formulas~\ref{delibone} and~\ref{delibtwo} therefore
predict Sentences~\ref{wronga} and~\ref{wronge}, along with many other
strange possibilities, are correct.

\ex. \label{wronga} John deliberately Mary hit.

\ex. \label{wronge} Mary the friend of deliberately left.

Other higher-order Lambek calculus formulas have similar problems when
we try to translate them into ACG. For example, the word ``and'' when
used for the coordination of transitive verbs has 3024 possible
translations, with 420 generating the correct surface structure and
148 having, in addition, the correct semantics as a possible
reading. However, these many possibilities all follow the same pattern
we have seen above for adverbs: they use a combination of
extraction-like and quantifying-in constructions and therefore overgenerate.




\subsection{Reflexives in D and in first-order linear logic}

There is a minor difference between the Displacement calculus and its
translation into first-order linear logic. The object reflexive of
\citeasnoun{mvf11displacement}.

$$
((vp \upl np) \upr np) \downr (vp \upl np)
$$

\noindent is translated as follows, when it occurs at string positions
3-4 (the $vp$ subformulas have been left untranslated).
$$
\forall x_0,x_1,x_2,x_5 [  np(3,4) \multimap np(x_1,x_2)
\multimap \| vp \|^{x_0,x_5} ] \multimap np(x_1,x_2) \multimap
\| vp \|^{x_0,x_5}
$$

The first-order linear logic formula more or less states that it transforms
a ditransitive verb (with one of its objects occurring at position
3-4) into a transitive verb. However, the original D formula specifies
something more, namely that the other $np$ argument, $np(x_1,x_2)$
occurs before this first $np$, in other words that $x_2 \leq 3$. It is
this property which is used to ensure the following two grammaticality judgments.

\ex. Mary talked to John about himself.

\ex. $*$ Mary talked about himself to John.

For formulas of the form $(A\multimap B \multimap C)
\multimap D$, the Displacement calculus can encode the precedence
between the $A$ and $B$ subformulas, whereas its translation in
first-order linear logic cannot, in can only unify positions but has
no mechanism for making statements like $x_2 \leq 3$\footnote{It would be possible to enrich first-order
  linear logic with order constraints on the variables, specifying
  things like $x_1 \leq x_2$. Though such an extension would be
  simple to implement, especially in a constraint-programming
  context, it seems to be rather
  powerful and it would complicate the proof theory, since such
  constraints would in all likelihood need to be non-linear.}.

Though I think the application of this property to reflexives is a
clever solution, I am unsure if its unavailability in first-order
linear logic is truly a handicap. The
Displacement calculus refers to its ``separators'' by number in the
linear order, whereas the lambda calculus-like formalisms give their
``points for future insertion'' a name (ie.\ a lambda calculus
variable). Exploiting the core mechanism of the Displacement calculus
to enforce constraints on linear order is a delicate undertaking since
it risks interfering with the many other operations depending on these
separators. For the treatment of Dutch verb clusters,
\citet[p.\ 28]{mvf11displacement} are already aware that they
sometimes need to refer to separators by mechanisms other than linear
order.

\subsection{Open questions}

One important question must remain unanswered here: can we give
(partial) translations from the Displacement calculus to Hybrid
Type-Logical Grammars, or vice versa? We can, of course, translate
into first-order linear logic and see if can translate it back to the
other system \citep[uses such a strategy, but to translate first-order
linear logic proofs back to proofs in the ``source
language'']{moot15l1}. However, such a strategy depends
on the exact formula obtained, for example the Formulas~\ref{f:d} and~\ref{f:lambda}
on page~\pageref{f:d} are equivalent and therefore this elementary
case would already need some additional mechanism to cope with equivalence.
A direct translation might side-step this problem and help crystallise
the differences between the systems even more clearly.

There is another open question, which was already briefly alluded to
when discussing unification. The variables and quantifiers of the
formulas which interest us follow specific patterns. For example, each
quantifier binds two occurrences of its eigenvariable (this is true
for position variables such as those we have seen here, for agreement
variables, such as case, grammatical gender etc., we may allow quantifiers which bind only one
occurrence of their eigenvariable). Position variables further have
one positive and one negative occurrence (where the switch from left
position to right position is treated as a polarity change). Are there
further patterns to discover here?

\section{Towards convergence}

\begin{figure}
\begin{center}
\includegraphics[width=9cm]{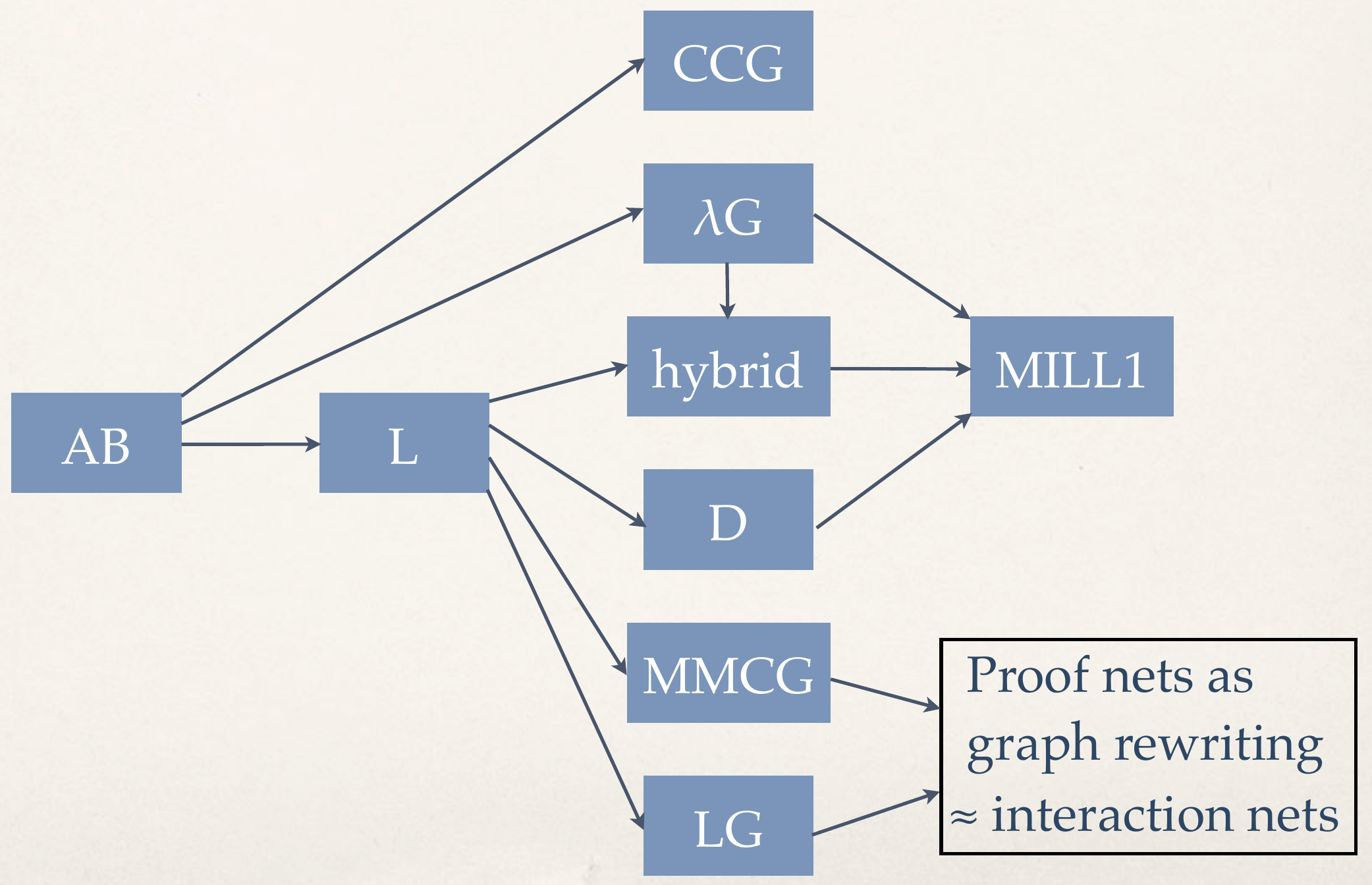}
\caption{A (slightly simplified) representation of principal extensions and variants of the Lambek calculus
  (L). The Ajdukiewicz/Bar-Hillel calculus (AB), Combinatory
  Categorial Grammars (CCG), lambda grammars ($\lambda G$), Hybrid
  Type-Logical Grammars (hybrid), the Displacement calculus (D),
  multimodal categorial grammars (MMCG) and the Lambek-Grishin (LG) calculus
and the two meta-logics: first-order linear logic (MILL1) and the proof
nets of Moot \& Puite}
\end{center}
\label{fig:meta}
\end{figure}

Figure~\ref{fig:meta} gives a slightly simplified picture of the main logical calculi and their relations.
The two meta-logics are not exclusive: the Displacement calculus can
be embedded both in a graph rewriting calculus (as shown by
\cite{ov14}) and in first-order linear logic
\cite{moot13lambek}. Similarly, $\textrm{NL}_{\lambda}$ 
seems to have a representation in both systems
\cite{bs14cont,moot13lambek,hal-00996724}, though the full formal
details of this possibility still need to be worked out. However, the
connection between a multimodal setup and a lambda calculus-like setup
explored by \citet[Chapter~17]{bs14cont} seems an important step towards
convergence of the meta-logics discussed in this paper.

It remains an open question
whether there is a natural way to guarantee convergence of the two
logics: what restriction on graph rewriting for proof nets would
ensure that the calculus can be translated in first-order linear
logic?

 Similarly, it is unclear at the moment whether there is a connection between the Lambek-Grishin calculus
(LG) and first-order linear logic. The multi-conclusioned nature of
the Lambek-Grishin calculus makes it hard to compare it directly to its
intuitionistic peers.

\section{Conclusions}

In this paper I have given two tools which I hope will create bridges
between the many formalisms in the family of type-logical
grammars. These tools are not only theoretical tools, but they are also
the basis of two implementations \cite{moot15grail0,moot15l1}, a
theorem prover based on (for the moment only the intuitionistic part
of) the graph rewriting calculus and a
theorem prover for first-order linear logic, which can output proofs
for the Displacement calculus and Hybrid Type-Logical Grammars by translation.

I believe open discussion of the benefits and disadvantages of one
system over another as well as a growing body of linguistic data which
can find a satisfactory account in type-logical grammars will be an
important factor in ensuring that our community stays vibrant and healthy.

\section*{Acknowledgements}

I would like to thank Chris Barker, Philippe de Groote, Yusuke Kubota, Robert Levine and
Michael Moortgat for their discussion on the topics which led to the
publication of this paper.

I would also like the referees of the Empirical Advances in Categorial
Grammar (2015) workshop for the remarks and comments on the submitted abstract.

\bibliographystyle{cg2015}
\bibliography{moot}

\end{document}